\documentclass[runningheads]{llncs}

\usepackage{cite}
\usepackage{amsmath,amssymb,amsfonts}
\usepackage{graphicx}
\usepackage{textcomp}
\usepackage{xcolor}
\def\BibTeX{{\rm B\kern-.05em{\sc i\kern-.025em b}\kern-.08em
    T\kern-.1667em\lower.7ex\hbox{E}\kern-.125emX}}
\usepackage{graphicx}
\usepackage{textcomp}
\usepackage{xcolor}
\usepackage{algorithm}
\usepackage[noend]{algpseudocode}
\usepackage{graphicx}
\usepackage{xcolor}

\usepackage{graphicx}
\usepackage[T1]{fontenc}
\usepackage[utf8]{inputenc}
\usepackage{textcomp}
\usepackage{amssymb,amsmath,amsfonts}
\usepackage{algorithm,algpseudocode}
\usepackage{etoolbox} 
\usepackage{booktabs}
\usepackage{graphicx}
\usepackage[hang, labelfont=bf,textfont=it]{caption}

\usepackage{algorithm}

\usepackage{graphicx}
\usepackage{xcolor}
\usepackage[hyphens,spaces]{url}

\begin{document}
\title{Iterative Mask Filling: An Effective Text Augmentation Method Using Masked Language Modeling
}
%
%
\author{Himmet Toprak Kesgin \orcidID{0000-0001-7554-1387} \and
Mehmet Fatih Amasyali \orcidID{0000-0002-0404-5973}}
\authorrunning{HT Kesgin, MF Amasyali}
%
\institute{\textit{Department of Computer Engineering} \\
\textit{Yildiz Technical University} Istanbul, Turkey \\}
%
\titlerunning{Iterative Mask Filling}
\maketitle              
\begin{abstract}
Data augmentation is an effective technique for improving the performance of machine learning models. However, it has not been explored as extensively in natural language processing (NLP) as it has in computer vision. In this paper, we propose a novel text augmentation method that leverages the Fill-Mask feature of the transformer-based BERT model. Our method involves iteratively masking words in a sentence and replacing them with language model predictions. We have tested our proposed method on various NLP tasks and found it to be effective in many cases. Our results are presented along with a comparison to existing augmentation methods. Experimental results show that our proposed method significantly improves performance, especially on topic classification datasets.

\end{abstract}
\section{Introduction}
Training neural networks with larger amounts of data can improve their generalization ability and performance. In fact, increasing the amount of data often has a greater impact on model performance than using a more complex model \cite{brants2007large}. However, obtaining large quantities of data can be expensive, especially in supervised learning, where each sample must be labeled.

Data augmentation is a way to increase the amount of training data available without collecting and labeling more data. In machine learning, data augmentation involves generating synthetic data from existing data in order to increase the amount of training data. Data augmentation can serve as a regularizer and improve the performance of machine learning models.

In image datasets, data augmentation involves applying transformations such as rotating, cropping, and changing the brightness of images.
Data augmentation has been studied more extensively in computer vision than in natural language processing \cite{shorten2021text}.
Numerous studies have demonstrated the remarkable success of data augmentation on image datasets \cite{perez2017effectiveness, mikolajczyk2018data, fawzi2016adaptive}.
One reason for this; transformations create new, valid images when applied to data.
Applying transformations to text datasets is not straightforward because they can disrupt syntax, grammatical correctness, and even alter the meaning of the original text. In addition, it is more challenging to preserve the label of an augmented sample in text than in images. These difficulties make it challenging to find effective data augmentation methods for text datasets.

Existing text augmentation methods include back-translation\cite{hayashi2018back}, synonym word substitution\cite{rizos2019augment}, easy data augmentation (EDA) \cite{wei2019eda} techniques such as random insertion, random swap, random deletion, or transformer-based text generation techniques such as; GPT-3\cite{brown2020language} or BERT \cite{devlin2018bert}.

In this paper, we propose a novel augmentation method that leverages the Fill-Mask feature of the transformer-based BERT model. We have tested our proposed method on a variety of natural language processing (NLP) tasks and found it to be effective in many cases. We compare our proposed method to existing text augmentation methods and present our results. The rest of this paper is organized as follows: Literature review in section 2, methods in section 3, experiments in section 4, discussion and limitations in section 5, and conclusions in section 6.

\section{Literature Review}

Synonym replacement is one of the automated text augmentation techniques.
Synonym replacement is the process of replacing, especially, nouns or verbs with their synonyms from a formal database source.
Zhang et al. used the WordNet database for synonym replacement to increase the size of the training dataset \cite{zhang2015character}.

The EDA offers four simple data augmentation operations: random swaps, random insertions, random deletions, and synonym replacements \cite{wei2019eda}.
EDA methods are tested with RNNs and CNNs extensively in their experiments for text classification.
Authors found that model performance is significantly improved by EDA techniques, especially for small datasets.

As an alternative to EDA, AEDA (An Easier Data Augmentation) only incorporates random punctuation marks into the original text\cite{karimi2021aeda}.
The AEDA method is simpler to implement, and it preserves all the input sentence information, since it does not include deleting or replacing.
AEDA has shown to improve performance on 5 text classification datasets.

Word embeddings can be used for a similar purpose as well.
Instead of replacing words with synonyms from a specific source, With Word2Vec\cite{mikolov2013efficient} words are replaced with their most similar counterparts \cite{marivate2020improving}.
The authors demonstrate that when a formal synonym model is not available, Word2vec-based augmentation can be beneficial.

Text augmentation can also be accomplished using back-translation.
Back translation is the translation of a sentence from one language into another and then translating it back into the original.
It is shown that the use of back translation resulted in a decrease in overfitting and an improvement over the BLUE score for IWSLT tasks \cite{sennrich2015improving}.

There have been significant gains across different NLP tasks using transformer-based models, such as BERT\cite{devlin2018bert}, GPT-2\cite{radford2019language}, and BART\cite{lewis2019bart}.
It is possible to use the text generation capabilities of these models for text augmentation \cite{kumar2020data, coulombe2018text}.
The main challenge of these methods is preserving the label of the augmented sentence.

The augmentation of data does not have to be at the textual level.
Once the text is represented as a vector, various techniques such as noise injection \cite{xie2017data} or mix-up\cite{zhang2017mixup, sun2020mixup} can be applied for augmentation.
These methods are not specific to text augmentation, but can be used for other data types.
These augmentation methods can be applied both with and without text-level augmentation, yet they are not the main topic of our analysis.

We propose an augmentation method based on a transformer-based BERT model.
As opposed to the previous methods, it uses the masked language modeling (Fill-Mask) feature instead of directly generating text.
The Fill-Mask task involves masking and predicting which words to replace the masks.
These models are useful for obtaining statistical understanding of the language in which the model was trained.
A detailed description of our proposed augmentation method is can be found in the method section.

\section{Methods}
Increasing the number of samples in the training set can generally improve the performance of a machine learning model, but the extent of the improvement depends on how different the new samples are from the existing ones, whether they contain noise, and how accurate their labeling is. Adding new examples that are identical to existing ones may not have a significant impact on model training. On the other hand, if the new samples are sufficiently different from the existing ones, the model may be able to learn better decision boundaries. The performance of the model can also be negatively affected by a large amount of input or label noise in new samples.

In data augmentation, it is important that the generated instances are differentiated from the existing ones while still maintaining their labels. We propose Iterative Mask Filling Augmentation, a method that aims to replace existing sentences or paragraphs as much as possible while preserving their meaning and structure. This method uses masked-language modeling (MLM), which has been trained on a large corpus of text. MLMs predict words that have been intentionally hidden within a sentence and provide valuable information for understanding the statistical properties of language. Words can have different meanings in different contexts, so it is important to learn context-dependent representations of each word. In MLM, the context of a sentence is taken into account when generating mask predictions, allowing the model to make confident predictions about which words can replace masked words.

Iterative Mask Fill Augmentation is given in Algorithm 1. In this algorithm, each word in a sentence is replaced with the <Mask> symbol. The MLM model then determines which words can replace the masked word, along with their associated scores. The k hyperparameter of the algorithm determines how many words the new word is chosen from. There is a confidence score associated with each word that can replace the <Mask>, and these scores are normalized to probability values such that their sum is one. The word to be replaced is selected based on these probabilities.

Since the MLM model is trained on a large corpus of text, it suggests only plausible words for the given context. Taking into account the context of the entire sentence, if only one word seems plausible at a given point, it is most likely to be selected due to its high score in comparison to other words. Therefore, not all the words in the sentence will be changed; only those that are reasonable to change will be replaced. In cases where there is a chance of more than one word being plausible for a given point, probabilistic selection is performed, and the word is replaced. At the end of each iteration, an augmented version of the input sentence is created.

\begin{algorithm}
\caption{Iterative Mask Fill}
\begin{algorithmic}[1]
\Require
\State $MLM$ : Pretrained Mask Language Model
\State $k$ : Number of top labels to be returned by MLM
\State $Sent$ : Sentence to be augmented

\Statex

\Procedure {generate\_augmentation}{}
\State $tokenized\_sent \leftarrow \Call{word\_tokenize}{Sent} $
\State $l \leftarrow \Call{len}{tokenized\_sent} $

\For {$i \in \{1...l\}$}
\State  $tokenized\_sent[i] \leftarrow "<mask>" $
\State $ scores, preds \leftarrow \Call{MLM}{tokenized\_sent[i]}$
\State $word \leftarrow \Call{select\_word}{scores, preds, k} $
\State  $tokenized\_sent[i] \leftarrow word $
\EndFor

\State \Return $ \Call{Join}{tokenized\_sent} $

\EndProcedure

\Procedure {select\_word}{scores, preds, k}
\State $ sum = \Call{sum}{scores}$
\State $ p = scores / sum$
\State $j$ $\xleftarrow{}$ Choose $j$ from [1…k] with probability p


\State \Return $ preds[j] $
\EndProcedure

\end{algorithmic}
\end{algorithm}

For example, the augmented version of the sentence "We introduce a new language representation model called BERT" produced by the algorithm is "they developed a natural language processing system called BERT".
The steps of the algorithm can be seen in Figure 1. 

\begin{figure}[!h]
\begin{center}
\includegraphics[width=\columnwidth]{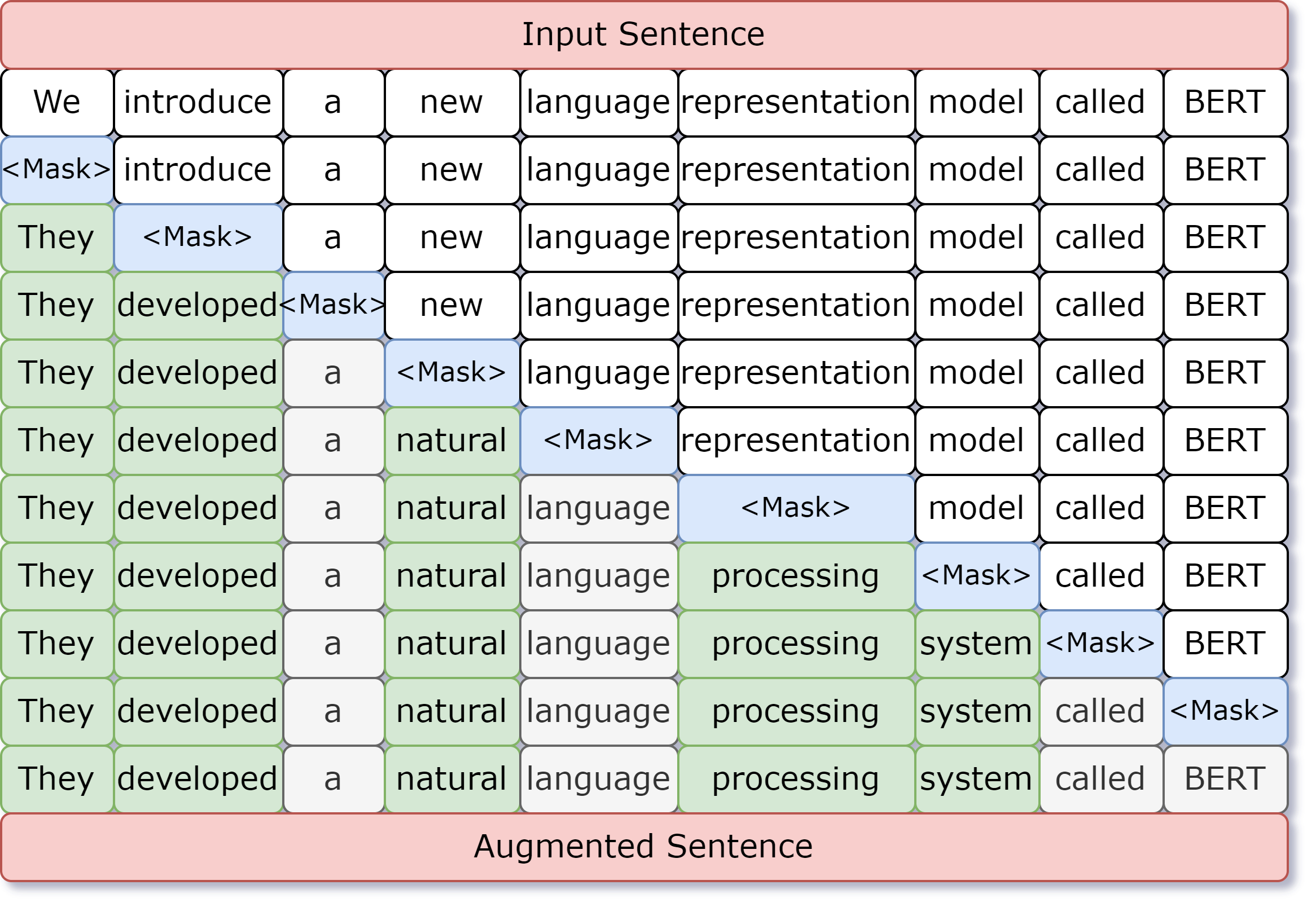}
\caption{Example Sentence Augmentation}
\end{center}
\end{figure}

\section{Experiments}

In this section, we will introduce the datasets used, describe the details of the experiments we performed, compare augmentation algorithms, and propose several improvements.

\subsection{Datasets}

We determined text data sets with various tasks for experiments. 
These datasets are the news category dataset (News) \cite{misra2018news, NewsCategoryDatasetKaggle}, financial sentiment analysis dataset (FinSent) \cite{malo2014good}, twitter sentiment analysis dataset (TwitSent) \cite{TwitterSentimentAnalysisKaggle} and the New York Times news (New York).
The New York dataset contains 800 news articles for training and 3000 news articles for testing. The News dataset has 10 classes, the New York dataset has 4 classes, and the FinSent and TwitSent datasets have 3 classes each. The datasets News, New York, TwitSent, and FinSent consist of 45000, 800, 74664, and 5843 samples, respectively, although we used subsets of these datasets in our experiments (as shown in Figure 2).

We created the New York dataset with news articles from the year 2022, sourced from the New York Times website. We did this because existing language models may have been trained on older datasets, which could bias their predictions. However, since these news articles were written after the language models were created, they were not trained on these texts.

The datasets used in our experiments can be grouped into two broad categories: category determination and sentiment analysis. To examine the effects of real training examples on the training dataset, we split the training dataset into subsets of various sizes and tested the performance of the model for each subset size. For each dataset, we used the same test set in all experiments. Figure 2 displays the results of these analyses. By examining the differences between the model's performance on the real training samples and the augmented samples, we can calculate the accuracy gain that would be achieved by adding real training samples to the training dataset. The orange line in Figure 2 indicates the number of training samples chosen for the experiments. Since augmented sentences cannot be as good as real training examples, these differences represent the potential maximum success of augmentation methods. These analyses allow us to compare different augmentation methods and see their potential gains.

\begin{figure}[!h]
\begin{center}
\includegraphics[width=\columnwidth]{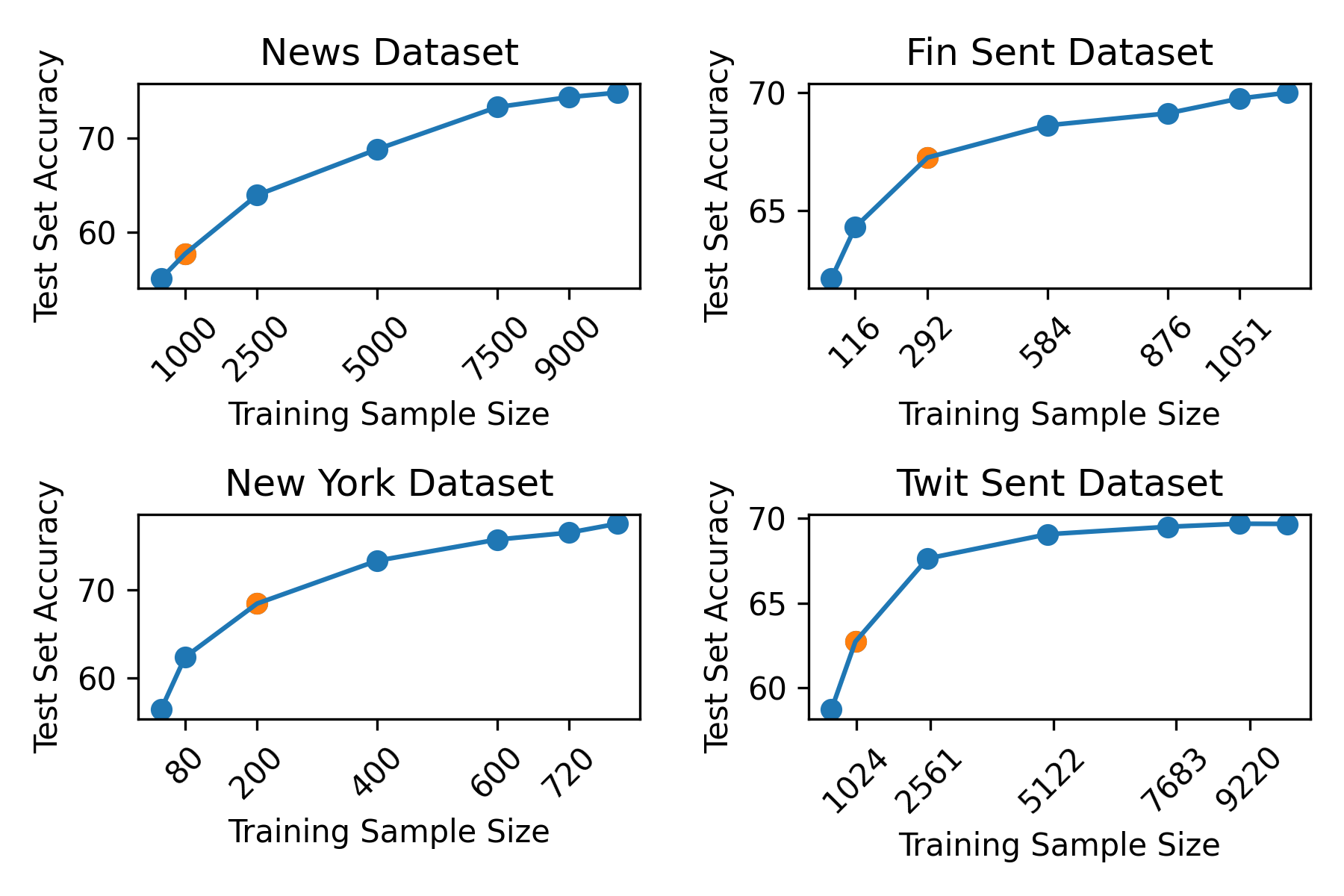}
\caption{Training Set Size - Test Set Accuracy Analysis}
\end{center}
\end{figure}

\subsection{Training Settings}
For classification, all texts in the dataset were converted to lowercase and sentence vectors were created using the transformer-based model\cite{sentencetransformer}.
Throughout all experiments, neural networks were used as classifiers.
First layer of the neural network consist of 384 neurons which are representations of texts.
Three hidden layers, consisting of 64, 16, and 4 neurons respectively, are then connected to this layer.
Tanh is the activation function of these layers.
Output layer of neural network determined by datasets number of classes.
The activation function of the last layer is softmax, which gives confidence scores for each class.

In all experiments, we used a hyperparameter $k$ as 5 in the Iterative Mask Fill (IMF) algorithm. This parameter determines how many words are considered as candidates to replace the masked word at each iteration of the IMF algorithm. The value of k can affect the performance of the IMF algorithm, and choosing an appropriate value for k may require experimentation. In general, increasing the value of k may allow the IMF algorithm to consider a wider range of words as candidates to replace the masked word, potentially leading to more diverse augmented sentences. However, a larger value of k may also lead to slower execution of the algorithm and may result in augmented sentences that are less faithful to the original text. In our experiments, we found that a value of $k = 5$ produced good results for the datasets and tasks considered. It is possible that different values of k may be more suitable for different types of data and tasks.

\subsection{Comparison of Text Augmentations}
The purpose of this section is to test how much the Iterative Mask Fill (IMF) augmentation algorithm improves performance by adding new examples to the training set. We compare the performance of the IMF Augmentation algorithm to several other basic text augmentation algorithms, including random insertion, random swap, random deletion, synonym word substitution, back translation augmentation, and BERT replacement.

Random insertion (ri): This method involves inserting one of the synonyms of a randomly chosen word anywhere in the sentence.
Random swap (rs): This method involves replacing two randomly chosen words in a sentence.
Random delete (rd): This method involves randomly deleting each word in a sentence based on a probability p.
Synonym word substitution: This method involves replacing n words with one of their synonyms among the non-stop words in a sentence.
Back translation (bt): This method involves translating a sentence from one language to another and back to the original language. In our experiments, the data sets we used were in English, so we translated the sentences into Turkish and augmented them by translating them back into English.
BERT replacement (br): This method involves masking some words in sentences and replacing them with BERT predictions. This method is similar to the non-iterative version of our proposed IMF method, as well as the synonym replacement method, where words are replaced with BERT predictions instead of their synonyms.
In all of these methods, we used an alpha ratio of 0.1, which yielded the best performance in our EDA study. The alpha ratio indicates the percentage of words in the sentence that will be changed. Unless it is 0, it guarantees that at least one word will be changed.

In this section, we compare the performance of various methods of text augmentation. Based on the training size-accuracy analysis shown in Figure 2, we used the training set sizes indicated by the orange line in the figure. For each sample, we included 1 and 4 augmentations using the existing methods. Therefore, the training set size increased by 100\% and 400\% in each set of experiments. As part of the performance comparisons, we also included the results when real sentences were added to the training dataset. The results of these experiments are shown in Table 1.

\begin{table}[!h]
\centering
\caption{Comparison of text augmentation methods}
\label{tab:my-table}
\resizebox{\columnwidth}{!}{%
\begin{tabular}{@{}|c|cccc|@{}}

\toprule
\multicolumn{1}{|c}{} & New York & News  & Fin Sent & Twit Sent \\ \midrule
vanilla              & $69.46 \pm 2.74$ & $58.98 \pm 2.19$ & $67.23 \pm 0.36$   & $62.25 \pm 0.64$    \\
real\_sample         & $74.70 \pm 1.71$ & $63.35 \pm 2.32$ & $68.59 \pm 0.53$   & $65.16 \pm 0.70$    \\
\midrule
ri (100\%)                   & $71.13 \pm 1.42$            & $60.30 \pm 1.56$  & $64.94 \pm 0.59$   & $ 61.73 \pm 0.54 $  \\
rs (100\%)                  & $69.50 \pm 2.37$              & $59.03 \pm 2.47$ & $64.73 \pm 0.75$   & $61.53 \pm 0.47$    \\
rd (100\%)                  & $ 70.61 \pm 2.45$            & $57.81 \pm 3.23$ & $64.66 \pm 0.62$   & $60.93 \pm 0.75$    \\
sr (100\%)                  & $69.76 \pm 2.06$             & $58.17 \pm 3.13$ & $65.06 \pm 0.59$   & $61.46 \pm 0.41$   \\
bt (100\%)                  & $69.58 \pm 2.38$             & $60.73 \pm 3.50$ & $66.81 \pm 1.34$   & $62.69 \pm 0.35$    \\
br (100\%)                  & $68.58 \pm 2.69$             & $58.40 \pm 1.47$ & $67.12 \pm 0.68$   & $61.82 \pm 0.46$   \\
imf (100\%)                 & $71.43 \pm 1.68$             & $60.67 \pm 1.92$ & $66.56 \pm 0.82$   & $61.75 \pm 0.44$  \\ \midrule
ri (400\%)                  & $69.05 \pm 2.09$             & $59.58 \pm 1.73$ & $64.90 \pm 0.78$   & $61.29 \pm 0.36$    \\
rs (400\%)                  & $69.92 \pm 2.35$             & $60.13 \pm 2.27$ & $64.73 \pm 0.66$   & $61.52 \pm 0.69$    \\
rd (400\%)                  & $70.41 \pm 1.88$             & $59.52 \pm 1.68$ & $64.63 \pm 0.75$   & $61.63 \pm 0.38$    \\
sr (400\%)                  & $70.69 \pm 1.38$             & $60.45 \pm 2.81$ & $65.36 \pm 0.68$   & $61.99 \pm 0.61$    \\
br (400\%)                  & $69.79 \pm 1.53$             & $\mathbf{61.52} \pm 1.71$ & $67.09 \pm 0.33$   & $61.76 \pm 0.56$    \\ 
imf (400\%)                 & $\mathbf{71.93} \pm 1.58$    & $\mathbf{61.25} \pm 1.64$ & $64.47 \pm 0.70$   & $60.77 \pm 0.78$   \\ \bottomrule
\end{tabular}%
}
\end{table}

These results show that the basic text augmentation methods generally improve the vanilla performance on category classification datasets, but not on sentiment analysis datasets. In sentiment analysis, the back translation method only improved performance on the TwitSent dataset. The EDA methods did not improve success much, but random insertion was the most promising among them. The IMF method improved performance on the two category classification datasets but decreased performance on the sentiment analysis datasets. Increasing the number of augmented sentences reduced overall accuracy, which suggests that the augmented sentences may contain label noise.

\subsection{Improving Performance of Text Augmentations}
While performing data augmentation, it is important to preserve the label of the instance. If the label changes during the augmentation process, it will mislead the model and reduce its performance. On the other hand, if the label is preserved without significantly altering the instance, the effect on the model's performance will be minimal, since the instance's representation in space will not be changed. Among the augmentation methods we used in the experiments, IMF is the one that changes the input sentence the most, since it has the potential to change each word in the sentence.

To visualize the impact of different augmentation methods on sentence representations, we plotted the vector representations of real and augmented texts using 2-dimensional TSNE. For each dataset, we randomly selected 100 representations and plotted the real and augmented sentences with different colors. Figure 3 illustrates these plots. As shown in Figure 3, the real and augmented sentences completely overlap for the other methods, but there are differences for the IMF method. These experiments support our claim that sentences augmented with IMF are more different from their original counterparts.

\begin{figure}[!h]
\begin{center}
\includegraphics[width=\columnwidth]{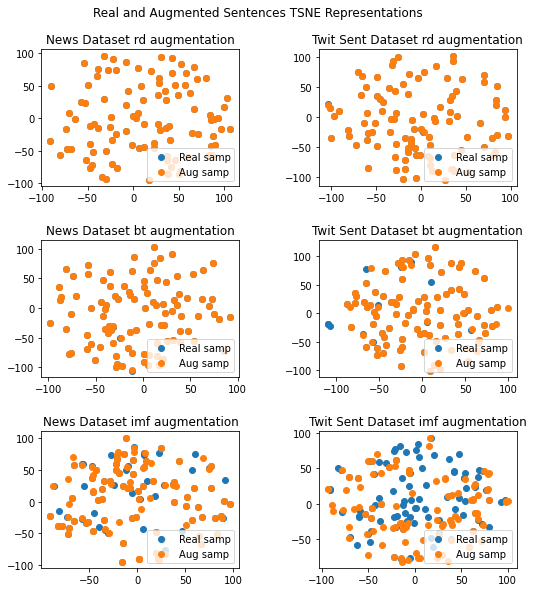}
\caption{Real and Augmented Sentences TSNE Representations}
\end{center}
\end{figure}

In this section, we also propose a method for filtering the texts generated by augmentation methods, selecting some of them, and including only the selected examples in the training set. To do this, we use a model trained with the vanilla method without augmentation for sample selection. With this model, we calculate the loss values of the augmented sentences and include only the k\% of augmented sentences with the lowest loss values in the training set. This method allows us to choose only the most realistic and useful augmented examples for training the model.

\begin{table}[!h]
\centering
\caption{Effect of filtering augmented sentences with low loss}
\label{tab:my-table}
\resizebox{\columnwidth}{!}{%
\begin{tabular}{@{}|cccccc|@{}}
\toprule
\multicolumn{1}{|l}{} & \% smallest loss                            & New York & News  & FinSent & TwitSent \\
vanilla              & \multicolumn{1}{l}{{\color[HTML]{333333} }} & $69.46 \pm 2.74$    & $58.98 \pm 2.19$ & $67.23 \pm 0.36$   & $62.25 \pm 0.64$    \\
real\_sample         & \multicolumn{1}{l}{}                        & $74.70 \pm 1.71$ & $63.35 \pm 2.32$ & $68.59 \pm 0.53$   & $65.16 \pm 0.70$    \\ \midrule
ri & 100 \%                  & $71.13 \pm 1.42$            & $60.30 \pm 1.56$  & $64.94 \pm 0.59$   & $ 61.73 \pm 0.54 $  \\
rs & 100 \%                  & $69.50 \pm 2.37$              & $59.03 \pm 2.47$ & $64.73 \pm 0.75$   & $61.53 \pm 0.47$    \\
rd & 100 \%                  & $ 70.61 \pm 2.45$            & $57.81 \pm 3.23$ & $64.66 \pm 0.62$   & $60.93 \pm 0.75$    \\
sr & 100 \%                 & $69.76 \pm 2.06$             & $58.17 \pm 3.13$ & $65.06 \pm 0.59$   & $61.46 \pm 0.41$   \\
bt & 100 \%                  & $69.58 \pm 2.38$             & $60.73 \pm 3.50$ & $66.81 \pm 1.34$   & $62.69 \pm 0.35$    \\
br & 100 \%                  & $69.79 \pm 1.53$             & $61.52 \pm 1.71$ & $67.09 \pm 0.33$   & $61.76 \pm 0.56$  \\
imf & 100 \%                 & $71.43 \pm 1.68$             & $60.67 \pm 1.92$ & $66.56 \pm 0.82$   & $61.75 \pm 0.44$  \\ \midrule
ri                   & 80 \%                                        & $71.98 \pm 2.62$    & $61.42 \pm 2.06$ & $66.48 \pm 0.21$   & $61.81 \pm 0.50$    \\
rs                   & 80 \%                                        & $72.80 \pm 0.51$    & $62.12 \pm 1.77$ & $66.06 \pm 0.38$   & $61.93 \pm 0.51$    \\
rd                   & 80 \%                                        & $71.90 \pm 1.58$    & $60.74 \pm 1.82$ & $66.28 \pm 0.35$   & $61.91 \pm 0.69$    \\
sr                   & 80 \%                                        & $71.01 \pm 2.29$    & $60.10 \pm 2.90$ & $67.08 \pm 0.29$   & $62.05 \pm 0.45$    \\
bt                   & 80 \%                                        & $69.64 \pm 2.29$    & $58.76 \pm 2.18$ & $66.88 \pm 0.54$   & $63.15 \pm 0.69$    \\
br                   & 80 \%                                        & $68.44 \pm 2.56$    & $62.25 \pm 1.21$ & $67.42 \pm 0.32$   & $62.27 \pm 0.40$    \\
imf                  & 80 \%                                        & $\mathbf{72.88} \pm 1.29$    & $\mathbf{62.87} \pm 2.06$ & $67.49 \pm 0.50$   & $\mathbf{63.24} \pm 0.50$    \\ \midrule
ri                   & 50 \%                                        & $69.44 \pm 2.21$    & $58.59 \pm 1.75$ & $66.67 \pm 0.74$   & $61.43 \pm 0.51$    \\
rs                   & 50 \%                                        & $66.71 \pm 3.23$    & $56.04 \pm 4.37$ & $66.97 \pm 0.37$   & $61.71 \pm 0.71$    \\
rd                   & 50 \%                                        & $68.19 \pm 2.08$    & $57.48 \pm 2.78$ & $66.89 \pm 0.83$   & $61.69 \pm 0.90$    \\
sr                   & 50 \%                                        & $70.20 \pm 2.25$    & $58.77 \pm 1.97$ & $67.27 \pm 0.31$   & $62.14 \pm 0.33$    \\
bt                   & 50 \%                                        & $71.04 \pm 1.29$    & $58.76 \pm 2.09$ & $67.08 \pm 0.34$   & $63.03 \pm 0.65$    \\
br                   & 50 \%                                        & $67.74 \pm 1.31$    & $55.77 \pm 3.41$ & $67.35 \pm 0.27$   & $62.11 \pm 0.39$    \\
imf                  & 50 \%                                        & $71.29 \pm 1.83$    & $62.07 \pm 1.63$ & $\mathbf{67.80} \pm 0.17$   & $62.03 \pm 0.52$    \\ \midrule
ri                   & 25 \%                                        & $68.80 \pm 2.59$    & $58.18 \pm 2.19$ & $66.32 \pm 0.96$   & $61.53 \pm 0.82$    \\
rs                   & 25 \%                                        & $68.80 \pm 3.67$    & $56.40 \pm 2.82$ & $66.70 \pm 0.47$   & $61.61 \pm 1.34$    \\
rd                   & 25 \%                                        & $68.92 \pm 2.40$    & $57.34 \pm 2.82$ & $66.93 \pm 0.48$   & $61.93 \pm 0.81$    \\
sr                   & 25 \%                                        & $68.45 \pm 3.55$    & $58.07 \pm 2.84$ & $66.31 \pm 1.54$   & $62.38 \pm 0.40$    \\
bt                   & 25 \%                                        & $70.00 \pm 3.05$    & $59.40 \pm 2.48$ & $67.47 \pm 0.37$   & $62.72 \pm 0.76$    \\
br                   & 25 \%                                        & $66.99 \pm 4.81$    & $57.45 \pm 1.45$ & $67.23 \pm 0.31$   & $61.68 \pm 0.84$    \\
imf                  & 25 \%                                        & $70.26 \pm 1.93$    & $59.33 \pm 1.44$ & $66.94 \pm 1.09$   & $61.36 \pm 0.69$    \\ \bottomrule
\end{tabular}%
}
\end{table}

Table 2 summarizes these results. In Table 2, four sentences were generated for each sentence in the dataset using every method except back translation, which only generates one augmented sentence for each sentence. For the lowest loss rates of 25\%, 50\%, and 80\%, we included the corresponding percentage of augmented sentences in the training set. These results show that including low-loss augmented sentences in the training set significantly improves the performance of the augmentation methods. We believe that low-loss examples are less likely to contain label noise. Filtering low-loss samples, particularly improved the performance of the IMF algorithm, which performed very closely to including real samples in the training set for category classification datasets. Across the datasets and algorithms, including the 80\% of augmented sentences with the lowest loss rates, except for FinSent, provided the best results.

\subsection{Using Different Language Models}

The IMF method we propose uses a masked language model to generate augmented sentences. For each sentence, it uses the predictions of the language model for the number of words in the sentence. As a result, the duration of the augmentation process depends on the number of words in the sentence and the prediction time of the model. Very large models, such as BERT, can take a long time to generate predictions because they have a large number of parameters. Therefore, in this section, we perform augmentation using smaller versions of BERT, namely distilBERT\cite{sanh2019distilbert} and tinyBERT\cite{turc2019well}, and compare their performance to the original BERT model. Table 3 summarizes these results.

\begin{table}[!h]
\centering
\caption{Comparison of different language models for the IMF}
\label{tab:my-table}
\resizebox{\columnwidth}{!}{%
\begin{tabular}{|ccccccc|}
\toprule
MLM        & param\_count & time & New York                     & News  & FinSent & TwitSent \\ \midrule
Bert      & 110m & 240 s & $72.88 \pm 1.29$ & $62.87 \pm 2.06$ & $67.49 \pm 0.50$   & $63.24 \pm 0.50$    \\
DistilBert & 66m & 156 s & $72.36 \pm 1.51$                        & $63.70 \pm 1.51$ & $66.85 \pm 0.50$   & $62.00 \pm 0.39$    \\
TinyBert  & 4m & 23 s & $72.38 \pm 1.63$                        & $61.25 \pm 2.10$ & $66.50 \pm 0.38$   & $61.88 \pm 0.52$    \\
\bottomrule
\end{tabular}%
}
\end{table}

In Table 3, the param\_count column indicates the number of parameters in the model, and the time column shows the time in seconds required for 100 sentence augmentations in the News dataset. In general, as the number of parameters in the language model decreases, the performance of the augmented sentences also decreases. However, reducing the size of the language model significantly speeds up mask estimation. This trade-off between performance and speed can be taken into consideration when selecting a language model for text augmentation. There is a close to linear relationship between the number of model parameters and the time required for mask estimation.

\section{Discussion and Limitations}

Previous methods proposed for text augmentation aim to enrich the dataset by making simple changes to the text. These methods are not very effective at improving model performance. The IMF method that we propose can increase the performance of augmented sentences more than other simple methods, but it requires a language model and can take a long time to generate augmented sentences, especially for longer texts such as paragraphs. As mentioned in the previous section, the algorithm can be accelerated by using a language model with fewer parameters. In our experiments, we found that the IMF method is more suitable for sentence-based augmentation and is more effective for news texts than for sentiment analysis datasets. This is because the sentiment of a sentence can change completely when a single word is changed, while the meaning of a news text is less sensitive to changes in individual words. Additionally, for all augmentation methods, as the size of the real dataset increases, the performance obtained with augmented texts decreases. This is because even when using real data, model performance changes very little as the dataset size increases. Therefore, augmentation methods are most effective when the dataset size is small.

\section{Conclusions}

In this paper, we propose a new text augmentation method using a MLM. We compare its performance with other augmentation methods on two news classification and two sentiment analysis datasets. Our results show that it can significantly improve the performance, especially in news classification datasets with a small number of training samples. We also propose a simple filtering process for augmented sentences to preserve their labels. We observe that existing augmentation methods do not significantly improve the performance of sentiment analysis tasks. One of the main challenges in text augmentation is to sufficiently modify the text while preserving its labeling. In future work, we aim to propose new augmentation methods using language models trained for different tasks that address these limitations. We also plan to study the effects of online text augmentation in future work.

\section*{Acknowledgment}

This study was supported by the Scientific and Technological Research Council of Turkey (TUBITAK) Grant No: 120E100.

\bibliography{citation}
\bibliographystyle{splncs04}

\end{document}